\documentclass[letterpaper,conference,10pt]{ieeeconf}  
\IEEEoverridecommandlockouts

\usepackage{amsmath,amsfonts,amssymb,bm}
\usepackage{graphics} 
\usepackage{epsfig} 
\usepackage{times} 
\usepackage{color}
\makeatletter
\let\NAT@parse\undefined
\makeatother
\usepackage[pagebackref]{hyperref}
\usepackage{xcolor}
\hypersetup{
    colorlinks=true,
    linkcolor={red!75!black},
    citecolor={blue!75!black},
    urlcolor={blue!75!black}
}
\usepackage{cite} 
\usepackage{url}
\usepackage{graphicx,epsfig,epstopdf,float} 
\usepackage{algorithm}
\usepackage[noend]{algpseudocode}
\usepackage{framed}
\usepackage{cancel}
\usepackage[normalem]{ulem}
\usepackage{diagbox}
\usepackage{pifont} 
\usepackage{listings, multicol}
\usepackage{textcomp} 
\usepackage{siunitx} 
\let\labelindent\relax
\usepackage{enumitem}
\usepackage{booktabs}  
\usepackage{subcaption}
\usepackage{multirow}

\definecolor{darkgreen}{RGB}{0,127,0}
\definecolor{darkred}{RGB}{200,0,0}
\definecolor{greentxt}{RGB}{106,168,79}
\definecolor{bluetxt}{RGB}{74,134,232}
\definecolor{redtxt}{RGB}{152,0,0}

\def\greencheckmark{\textcolor{darkgreen}{\checkmark}}
\def\redxmark{\textcolor{darkred}{\ding{55}}}  

\newcommand{\R}{\mathbb{R}}  

\title{\LARGE \bf Mitigating Covariate Shift in Imitation Learning for Autonomous Vehicles Using Latent Space Generative World Models}

\author{Alexander Popov, Alperen Degirmenci, David Wehr, Shashank Hegde, Ryan Oldja, Alexey Kamenev \\ Bertrand Douillard, David Nist{\'e}r, Urs Muller, Ruchi Bhargava, Stan Birchfield, Nikolai Smolyanskiy  \\
NVIDIA 
}

\begin{document}

\maketitle

\begin{abstract}
We propose the use of latent space generative world models to address the covariate shift problem in autonomous driving.
A world model is a neural network capable of predicting an agent's next state given past states and actions.
By leveraging a world model during training, the driving policy effectively mitigates covariate shift without requiring an excessive amount of training data. 
During end-to-end training, our policy learns how to recover from errors by aligning with states observed in human demonstrations, so that at runtime it can recover from perturbations outside the training distribution. 
Additionally, we introduce a novel transformer-based perception encoder that employs multi-view cross-attention and a learned scene query. 
We present qualitative and quantitative results, demonstrating significant improvements upon prior state of the art in closed-loop testing in the CARLA simulator, as well as showing the ability to handle perturbations in both CARLA and NVIDIA's DRIVE Sim.\footnote{Video is at \url{https://youtu.be/7m3bXzlVQvU}.}

\end{abstract}


\section{INTRODUCTION}

Autonomous vehicles must navigate complex and dynamic environments. In terms of both data and compute, the most effective method is to learn from human driving. Recent studies have trained neural planners using imitation learning (IL) \cite{barto1983imlearning}. However, IL is susceptible to the covariate shift problem \cite{ross2011reduction}, which impedes the development of effective driving policies.

Covariate shift occurs when the state distribution encountered by the planner's policy during deployment differs from that during training. This discrepancy arises because the training data, which captures expert behavior, may not encompass all possible states the policy will face in practice. As a result, the driving policy may perform well on the training data but fail in new, unseen states when deployed. This results in the autonomous vehicle drifting away from optimal trajectories when guided by the trained neural planner---oftentimes with catastrophic effects.

One way to mitigate covariate shift is to collect recovery trajectories from bad states to good states via new human demonstrations.  Another way is to augment data by stochastically sampling new states and then using a privileged planner in simulation to compute recovery trajectories \cite{ross2011reduction, chang2021mitigating}. Unfortunately, such methods are fragile, typically use sub-optimal heuristics, and are expensive to implement.

In this paper, we demonstrate that co-training driving policies with generative latent space world models mitigates the covariate shift problem. A world model, as described in \cite{WorldModels:Ha:2018}, is a deep neural network (DNN) trained to represent the ego vehicle’s state, traffic dynamics, and the static structure of the environment. Such models are trained using expert demonstrations recorded from driving data (including sensor data, vehicle trajectories, and navigation goals) to predict future states based on past states and driving actions.

\begin{figure}[!t]
    \centering
    \mbox{
      \parbox{0.98\columnwidth}{
        \centering
        \includegraphics[width=0.98\columnwidth]{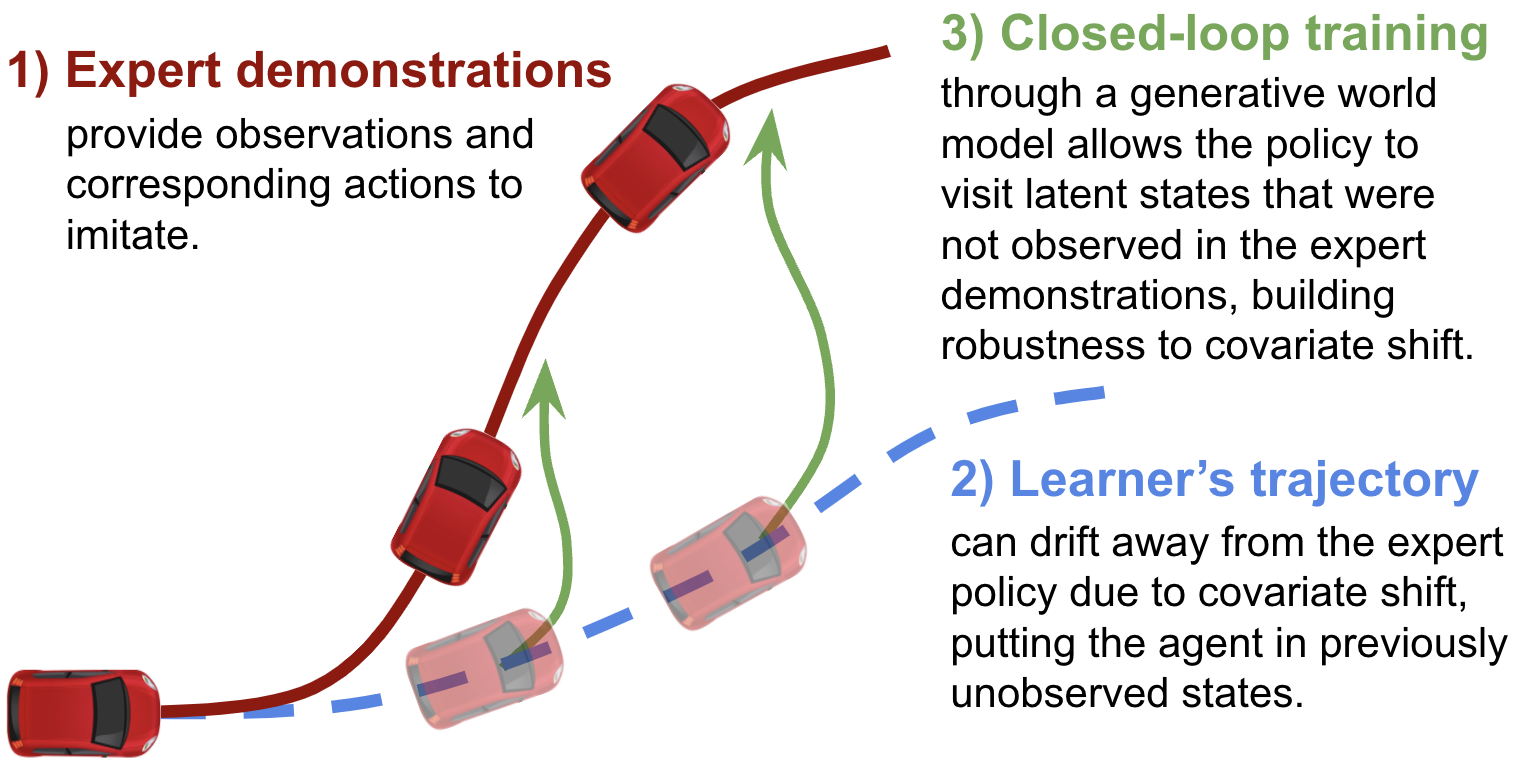}
      }
    }
    \caption{Training with a latent generative world model allows our policy to visit latent states that were not observed in the expert demonstrations, and learn to recover by minimizing the KL-divergence between world model rollouts and expert demonstrations, teaching the policy to select actions that guide the agent toward favorable states.}
\label{fig:covariate_shift_mitigation}
\end{figure}

We train a latent space generative world model that allows us to sample new ego states from the learned latent space that were not present in the training data. These sampled states are then used to train the driving policy to recover from errors, where the policy learns to pick actions such that future latent states are closer to the states observed in human demonstrations (Fig.~\ref{fig:covariate_shift_mitigation}). Our experiments show that using latent space generative world models can solve or mitigate the covariate shift problem in imitation learning.

Our contributions are as follows:
\begin{itemize}[leftmargin=*]
    \item We train an end-to-end driving network with a latent space generative world model that addresses the covariate shift problem.
    \item We introduce a novel transformer-based perception encoder architecture (leveraging DINOv2~\cite{oquab2023DINOv2} as a featurizer) that facilitates the learning of effective world state representations used by the world model. 
    \item We present qualitative and quantitative results of closed-loop driving in both the CARLA simulator \cite{dosovitskiy2017carla} and the more realistic NVIDIA DRIVE~Sim simulator.
\end{itemize}

\section{PREVIOUS WORK}

\subsection{Covariate Shift in Imitation Learning}


Ross et al. \cite{ross2011reduction} introduce DAgger that structures prediction problems as no-regret online learning. This approach ensures performance improvements over time by iteratively aggregating novel states where the agent may fail through a training-deployment-collection-training loop. Chang et al. \cite{chang2021mitigating} address the covariate shift challenge with limited offline data using model-based imitation learning (MILO). MILO learns an effective policy with partial expert action coverage, enhancing robustness in real-world applications. However, it assumes direct access to the world state, does not consider noisy sensor state estimation, uses a discrete MDP, and assumes calibrated transition function uncertainty. Spencer et al. \cite{spencer2021feedback} argue that the divergence between “held out” error and performance of the learner \textit{in situ} is a manifestation of covariate shift, which can be mitigated by taking advantage of a simulator without further need for expert demonstrations. Ke et al. \cite{ke2021grasping} address the covariate shift in model-free imitation learning for manipulation tasks like grasping with chopsticks. Their technique enhances the robustness of learned policies, leading to improved performance in real-world scenarios. Tennenholtz et al. \cite{tennenholtz2021covariate} explore latent confounders’ covariate shift in both imitation and reinforcement learning (RL). The authors introduce techniques to detect and mitigate the impact of these confounders, enhancing the robustness of learned policies. In contrast to these approaches, we mitigate covariate shift by using the world model as a neural simulator to generate new states in the latent space during training.

\subsection{World Models}

Ha and Schmidhuber \cite{WorldModels:Ha:2018} introduced world model learning in 2018, training a model to forecast traffic in latent space, which could be decoded into Cartesian top-down space. They combined RNN and VAE to predict both top-down driving episodes and perspective views. The trained control policies using these world models allowed the policy to explore various scenarios and handle general cases as well as recover from unusual or adverse states. Dreamer~\cite{Dreamer:Hafner:ICLR2020} learns a latent space world model from a replay buffer of past experiences. It uses actor-critic RL and a learned world model to train a control policy. This system can learn long-horizon behaviors solely from images and has demonstrated state-of-the-art performance on various control tasks in the ATARI domain. DayDreamer \cite{DayDreamer:Wu:PMLR2023} is an extension of Dreamer applied to physical robot learning in the real world, without using simulators. MILE \cite{MILE:Hu:NEURIPS2022} adapted the Dreamer approach to the AV domain and tested it in the CARLA simulator \cite{dosovitskiy2017carla}. It learns a latent-space generative world model from expert driving demonstrations and uses it to train an end-to-end driving policy. Unlike Dreamer, it does not use RL in training and focuses on imitation learning. MILE can imagine future driving scenarios based on different ego actions, which are decoded as bird's-eye views (BEV). The system relies on LSS \cite{LSS:Phillion:ECCV2020} as its perception encoder. PredictionNet \cite{PredNet:Kamenev:ICRA2022} is an auto-regressive RNN-based world model that predicts future vehicle states, occupancies, and traffic flow for all traffic agents, including the ego. It can be used for ego action conditioned traffic prediction. Unlike Dreamer \cite{Dreamer:Hafner:ICLR2020}, it operates in a BEV space.

Our latent-space generative world model approach is similar to JEPA \cite{lecun2022jepa}, Dreamer \cite{Dreamer:Hafner:ICLR2020},  and MILE \cite{MILE:Hu:NEURIPS2022}. JEPA was only used for visual tasks, not vehicle planning and control. Unlike Dreamer, which employs a critic and RL for training, we derive our driving policy from human expert driving demonstrations, similar to MILE. Unlike MILE, we introduce a novel multi-view transformer-based perception encoder, utilize DINOv2~\cite{oquab2023DINOv2} as an image featurizer, use different training recipe and investigate covariate shift. 

\subsection{Planning and Control}

Several works proposed end-to-end approaches for AV using deep neural networks, including ChauffeurNet \cite{bansal2018chauffeurnet}, PlaNet \cite{hafner19planet}, and UniAD \cite{hu2023uniad}. These neural planners may encounter covariate shift, making it crucial to address this issue. ChauffeurNet mitigated it by introducing small random ego pose augmentations during DNN training. While this approach is similar to the concept in DAgger~\cite{ross2011reduction}, the specific set of augmentations is heuristically created, which is challenging to implement.

\begin{figure*}[t]
    \centering
    \mbox{
      \parbox{0.98\textwidth}{
        \centering
        \includegraphics[width=0.98\textwidth]{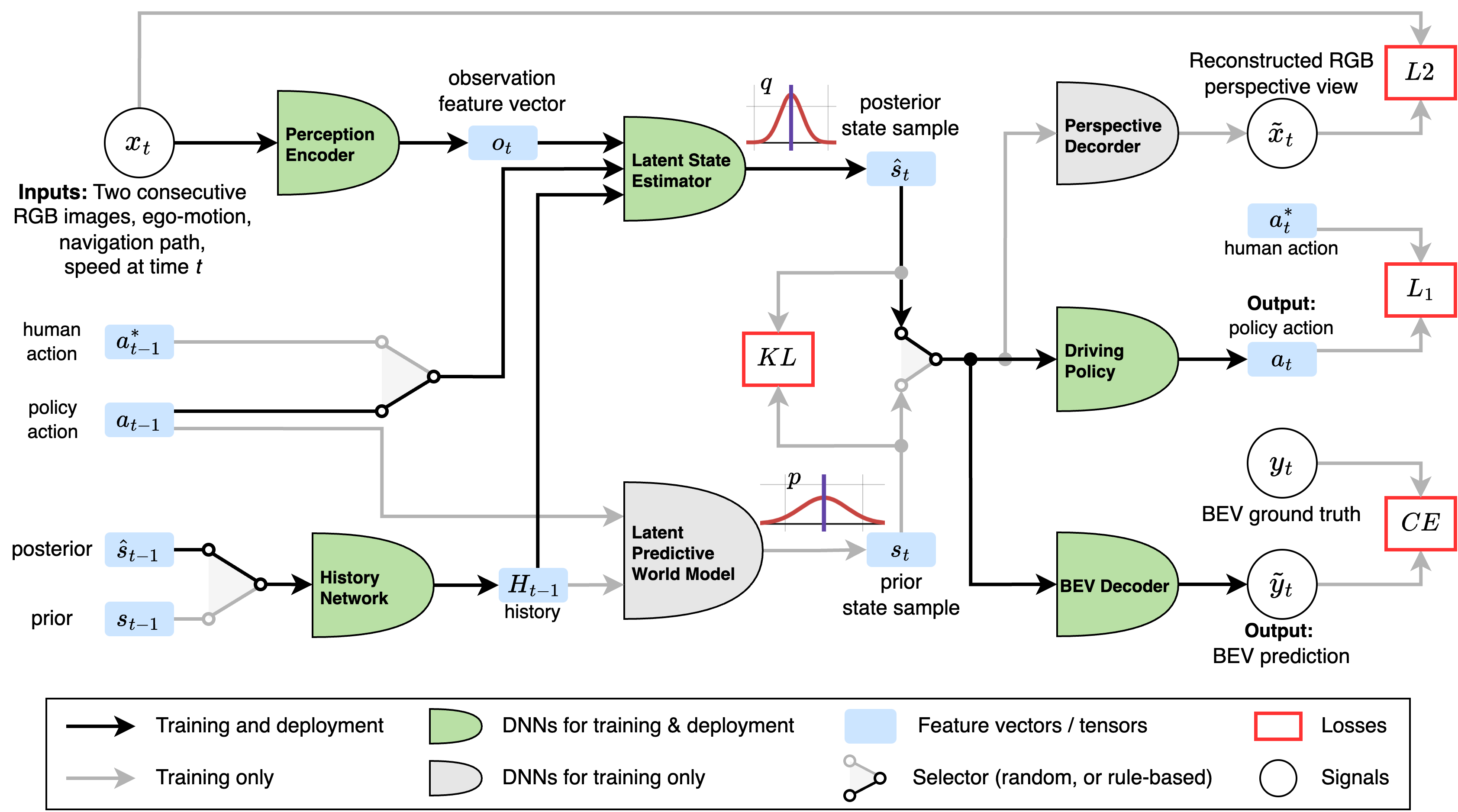}
      }
    }
    \caption{System architecture diagram showing transition from time $t-1$ to $t$. During training we backpropagate through an entire episode consisting of 12 timesteps (red boxes show the losses). During deployment the system is operated recurrently.}
    \label{fig:genwm_architecture}
\end{figure*}

\section{METHOD}
\label{sec:method}

\subsection{System Architecture}

Fig.~\ref{fig:genwm_architecture} illustrates the overall architecture of our system, comprising several DNNs co-trained end-to-end. 
Multilayer perceptrons (MLP) are used throughout the system for 
tensor shape adaptation between DNNs.

Intuitively, the architecture can be compared to behavior cloning (BC), where the perception encoder maps the input to an observation feature vector, which the driving policy uses to generate the next action. BC faces the covariate shift problem, since the policy is trained to follow the expert without being exposed to deviations. BC can be improved by tracking the state over time using a history network and state estimator, with backpropagation over the entire episode to capture long-term dependencies. However, covariate shift remains an issue when the system encounters unseen states.

To address covariate shift, our system uses a latent space generative world model that models state transitions conditioned on ego action. This transforms behavior cloning into a temporal world model-based imitation learning system. During backpropagation, gradients flow through the driving policy and world model, affecting previous actions over multiple time-steps, enabling recovery from mistakes. The recovery signal comes from the difference between latent states predicted by the world model and those computed by the state estimator. The system is trained end-to-end, with different networks converging at different rates, and the policy network converging last.

Why does this work?  The world model acts as a latent-space neural simulator that samples novel states at training time, exposing the system to these novel states to allow the policy to learn how to properly handle them. 
Since high-dimensional scene data are mapped to a lower-dimensional latent space with a smooth manifold, the trained policy is able to recover from rogue states (that is, move back toward states on the latent world model manifold that more closely resemble those associated with the expert demonstrations).

\subsection{System component details}

\begin{figure*}[t]
    \centering
    \mbox{
      \parbox{0.98\textwidth}{
        \centering
        \includegraphics[width=0.98\textwidth]{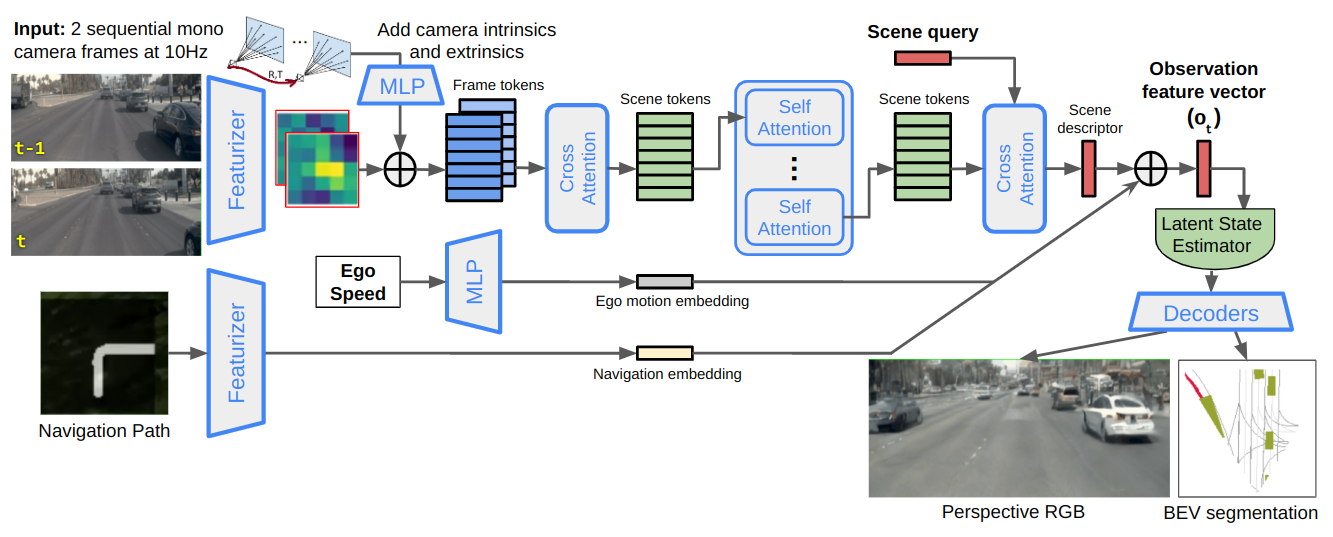}
      }
    }
    \caption{Our novel transformer-based multi-view perception encoder. Input data is tokenized, and frame tokens are cross-attended to match patches. Scene descriptor is computed via cross-attention between learned scene query and scene tokens. Output is computed by adding navigation and ego motion embeddings to the scene descriptor.}
    \label{fig:perception_encoder_diagram}
\end{figure*}

\textbf{Perception encoder} illustrated in Fig.~\ref{fig:perception_encoder_diagram} takes as input two sequential RGB frames (at \SI{10}{\hertz}) from the monocular front camera, corresponding past ego-motion (6~DOF vehicle pose transformation), the desired upcoming navigation path ($192 \times 192$ binary bird's-eye views (BEV) image, \SI{0.2}{\meter} per pixel resolution), and the current vehicle speed. 
Each RGB image is featurized by a DINOv2~\cite{oquab2023DINOv2} backbone to produce a set of image tokens $F_t$. Motion information is incorporated by a cross-attention layer \cite{vaswani2017} between queries $F_t$, keys, and values $F_{t-1}$ to produce a set of tokens $M_t$ that represent the currently observed scene. Camera intrinsics and extrinsics are processed by MLPs and added to the image features prior to cross-attention to help with multi-view matching similar to \cite{zisserman2021inputinductivebias}. $M_t$ is then processed via $4$ blocks of self-attention with $8$ heads into tokens $S_t$. Finally, a learned query $Q$ (similar to \cite{carion2020detr}) is used to cross-attend with keys and values to output a single vector to represent the observation feature vector $o_t \in \R^{512}$. The desired navigation path and ego speed are embedded via MLPs and then added to the scene embedding. This scene observation feature vector $o_t$ represents currently observed scene and is passed to the state estimator. Observation feature vector $o_t$ can also be concatenated with GPS and IMU data encoded as 1D vectors.

We use DINOv2 pre-trained backbone \cite{oquab2023DINOv2} as an image featurizer. This allows us to train a robust perception encoder on real data that can be deployed in a simulator for validation, without requiring fine-tuning to close the sim-to-real gap. If validation in simulation is successful, the trained end-to-end driving stack can be directly tested on a real vehicle, as it has been exclusively trained on real data.

\textbf{History network} aggregates previous states deterministically using the hidden state of a GRU RNN. We employ the same GRU settings as in DreamerV3~\cite{Dreamer:Hafner:ICLR2020}.

\textbf{State estimator} models the latent state posterior distribution conditioned on observations. It estimates a surrogate distribution $q({\hat s}_t | o_t, H_{t-1}, {\hat a}_{t-1})$ as a Gaussian, that is referred to as the posterior. It approximates the true hidden state using evidence lower bound (ELBO) \cite{blei2017varinf}. 
This generative DNN takes observed perception feature vector $o_t$, 
past history $H_{t-1}$ (which incorporates past state samples up to and including time $t-1$), featurized human expert ego action ${a^*}_{t-1} \in \R^{2}$ at time step $t-1$.
Expert ego action embedding is done by passing the combined steering and acceleration values via MLP. The state estimator computes the parameters of a Gaussian distribution, from which we then sample posterior latent states ${\hat s}_t \in \R^{512}$. This DNN is implemented as an MLP, as in DreamerV3~\cite{Dreamer:Hafner:ICLR2020}. The state samples are recursively used for the next iterations via the history network.

\textbf{World model} predicts stochastic latent space world transitions from $s_{t-1}$ to $s_{t} \in \R^{512}$ conditioned on ego action and history at time $t-1$. It takes as input the past history $H_{t-1}$, the ego action $a_{t-1} \in \R^{2}$ produced by the driving policy for time step $t-1$. The world predictor is a generative DNN that predicts the latent state as a Gaussian distribution, from which we sample the $s_t$ prior state $p(s_t | H_{t-1}, a_{t-1})$. The DNN is implemented as an MLP, similar to DreamerV3~\cite{Dreamer:Hafner:ICLR2020}. This DNN can be used to roll out new ``imagined" states without corresponding observations up to a given horizon. We can use these roll-outs for re-simulating long-tail situations. The latent predictive world model is not run at deployment time.

\textbf{Driving policy network} is an MLP network with four layers and ReLU activations. Input dimensions are the same as latent state dimensions and output dimension is two, corresponding to steering and acceleration/braking actions.

We iterate the system over time by using either $s_t$ or $\hat{s}_t$ for the next iteration with some probability. 
Switching between $s_t$ and $\hat{s}_t$ helps mitigate covariate shift problems. Since $s_t$ is sampled stochastically, it may include unusual states from which the model is trained to recover. This can be seen as a form of exploration (as in RL).

\textbf{The decoder DNNs} decode latent state samples $s_{(\cdot)}$ into perspective RGB views and bird's-eye views (BEV) semantic segmentation views. We experimented using GANs (StyleGAN2 \cite{karras2020stylegan2}) or latent diffusion model based on VideoLDM \cite{blattmann2023videoldm}\cite{Karras2022edm} with a pre-trained VAE from \cite{blattmann2023stable}, which worked much better than GANs. The decoders are trained using reconstruction losses against input sensor data (RGB frames) and top-down BEV ground truth targets. This allows decoding a sequence of latent states into a temporally cohesive perspective RGB sequence, which is useful for system inspection and visualization. The introspection capabilities provided by these decoders is critical for safety checks in end-to-end trained autonomy stacks.

\subsection{Training and Losses}

We use variational inference approach to train the system by maximizing the ELBO  \cite{blei2017varinf}.\footnote{Note that our ELBO-based loss is the same as the free-energy principle objective used in active inference within cognitive science \cite{friston2010fep}.}  Our objective function is ${\cal L} = {\cal L}_{Recon} + {\cal L}_{KL}$, similar to \cite{MILE:Hu:NEURIPS2022}, where 

\begin{align}
{\cal L}_{Recon} = \textstyle \sum_{t=1}^{T} \mathbb{E}_{p}
    \big[ & \,\,\,\,\,\, \log p \left(x_t \mid H_{t-1}, s_t \right) \nonumber \\
    &+ \log p \left( y_t \mid H_{t-1}, s_t \right) \nonumber \\
    &+ \log p \left( a_t \mid H_{t-1}, s_t \right) \big]
    \label{eq:one}
\end{align}
\begin{align}
\label{eq:kldivloss}
{\cal L}_{KL} = - \textstyle \sum_{t=1}^{T} \mathbb{E}_{q} \big[ D_{KL} &\big( q \left({\hat s}_t \mid o_t, H_{t-1}, {\hat a}_{t-1} \right) \| \nonumber \\
    &\  \, p \left(s_t \mid H_{t-1}, a_{t-1}\right) \big) \big]
\end{align}
where $x_t$ is the reconstructed RGB image, $y_t$ is the reconstructed BEV semantic label image, and $a_t$ is the action.  $p(x_t,\cdot)$ follows a Gaussian distribution, leading to L2 loss; $p(y_t, \cdot)$ follows a categorical distribution, leading to cross entropy loss; and $p(a_t, \cdot)$ follows a Laplace distribution, leading to L1 loss.

All DNN modules in our system were trained simultaneously by optimizing the objective $\cal L$. We have observed empirically that the training process proceeds in the following way: the perception encoder and decoder converge first, followed by the world predictor, and finally the driving policy.

\begin{table*}
 \caption{Results on CARLA closed-loop simulator 
 \cite{dosovitskiy2017carla} on held-out test data and new weather conditions using the same evaluation procedure as~\cite{MILE:Hu:NEURIPS2022}.  See text for details.
 }
 \label{tab:genwmnet_in_carla_scores}
 \begin{minipage}[t]{.95\linewidth}
 \centering
 \renewcommand{\arraystretch}{1.25}
 \begin{tabular}{lcccccc}
 \toprule
 \textbf{Method} & \textbf{Single Cam} & \textbf{LiDAR} & \textbf{Driving Score} $\uparrow$ & \textbf{Route} $\uparrow$ & \textbf{Infraction} $\uparrow$ \\ 
 \midrule
 CILRS \cite{codevilla2019exploring} & \textcolor{darkgreen}{yes} \greencheckmark & \textcolor{darkgreen}{no} \greencheckmark & 7.8 $\pm$ 0.3 & 10.3 $\pm$ 0.0 & 76.2 $\pm$ 0.5 \\ 
 LBC \cite{chen2020learning} & \textcolor{darkgreen}{yes} \greencheckmark & \textcolor{darkgreen}{no} \greencheckmark & 12.3 $\pm$ 2.0 & 31.9 $\pm$ 2.2 & 66.0 $\pm$ 1.7 \\ 
 TransFuser \cite{prakash2021transfuser} & \textcolor{darkred}{no} \redxmark & \textcolor{darkred}{yes} \redxmark & 31.0 $\pm$ 3.6 & 47.5 $\pm$ 5.3 & 76.8 $\pm$ 3.9 \\ 
 Roach \cite{zhang2021roach} & \textcolor{darkgreen}{yes} \greencheckmark & \textcolor{darkgreen}{no} \greencheckmark & 41.6 $\pm$ 1.8 & 96.4 $\pm$ 2.1 & 43.3 $\pm$ 2.8 \\ 
 LAV \cite{chen2022lav} & \textcolor{darkred}{no} \redxmark & \textcolor{darkred}{yes} \redxmark & 46.5 $\pm$ 3.0 & 69.8 $\pm$ 2.3 & 73.4 $\pm$ 2.2 \\ 
 TCP \cite{wu2022tcp} & \textcolor{darkgreen}{yes} \greencheckmark & \textcolor{darkgreen}{no} \greencheckmark & \emph{57.0} $\pm$ \emph{1.9} & \emph{85.3} $\pm$ \emph{1.2} & \emph{67.0} $\pm$ \emph{1.0} \\
 MILE \cite{MILE:Hu:NEURIPS2022} & \textcolor{darkgreen}{yes} \greencheckmark & \textcolor{darkgreen}{no} \greencheckmark & 61.1 $\pm$ 3.2 & 97.4 $\pm$ 0.8 & 63.0 $\pm$ 3.0 \\
 InterFuser$^*$ \cite{shao2022interfuser} & \textcolor{darkred}{no} \redxmark & \textcolor{darkred}{yes} \redxmark & \emph{68.3} $\pm$ \emph{1.9} & \emph{95.0} $\pm$ \emph{2.9} & --- \\
 TF++ \cite{jaeger2023tfpp} & \textcolor{darkgreen}{yes} \greencheckmark & \textcolor{darkred}{yes} \redxmark &  
 \emph{70.0} $\pm$ \emph{6.0} & \emph{99.0} $\pm$ \emph{0.0} & \emph{70.0} $\pm$ \emph{6.0} \\
 ReasonNet$^*$ \cite{shao2023reasonnet} & \textcolor{darkred}{no} \redxmark & \textcolor{darkred}{yes} \redxmark & \textbf{\emph{73.2}} $\pm$ \textbf{\emph{1.9}} & \emph{95.9} $\pm$ \emph{2.3} & \emph{76.0} $\pm$ \emph{3.0} \\ 
 \textbf{Ours} & \textcolor{darkgreen}{yes} \greencheckmark & \textcolor{darkgreen}{no} \greencheckmark & 70.0 $\pm$ 0.2 & \textbf{100.0} $\pm$ \textbf{0.1} & \textbf{80.0} $\pm$ \textbf{0.7} \\
 \midrule
 \multicolumn{2}{c}{Expert (RL agent with privileged info)~\cite{zhang2021roach}} && 88.4 $\pm$ 0.9 & 97.6 $\pm$ 1.2 & 90.5 $\pm$ 1.2 \\ \bottomrule
 \end{tabular}
 \end{minipage}
     \vspace{-0.05in}
\end{table*}



\section{EXPERIMENTS}

We conducted several sets of experiments to evaluate our system.
  
\subsection{CARLA Simulator Closed-loop Evaluation}

For this experiment, we trained our system using data from the CARLA  simulator~\cite{dosovitskiy2017carla}. We collected expert driving data in CARLA using a pre-trained agent (trained via reinforcement learning) following MILE~\cite{MILE:Hu:NEURIPS2022}. We gathered 50k episodes of driving, each 1.2 seconds long and consisting of 12 frames sampled at \SI{10}{\hertz} (to allow sufficient separation between frames). Our CARLA training dataset was collected in the CARLA simulator, similar to \cite{MILE:Hu:NEURIPS2022}, using a privileged RL-based expert driver~\cite{zhang2021roach}. We then jointly trained all the networks of our system end-to-end on this dataset. The training process spanned 15 million episodes, with an initial learning rate of $10^{-4}$ and an effective batch size of 64, utilizing the \emph{1cycle}~\cite{smith2018super} learning rate schedule. The total training time was 5 days on 16 NVIDIA A100 GPUs.

Our trained system was deployed in the closed-loop CARLA simulator \cite{dosovitskiy2017carla} for evaluation. Our system successfully reacted to traffic lights, followed lead vehicles, stopped for them, avoided collisions, allowed pedestrians to pass, and adhered to a given route. All training and inference were performed end-to-end without any handcrafted code. We quantitatively evaluated our system by computing closed-loop metrics according to the CARLA Leaderboard 1.0 (Sensors Track).\footnote{Like MILE~\cite{MILE:Hu:NEURIPS2022}, we were unable to submit to the online leaderboard, \url{https://leaderboard.carla.org/leaderboard} which is now closed.  CARLA leaderboard 2.0 was not possible, since it would require uploading our code; and it contains few methods.}  Following MILE~\cite{MILE:Hu:NEURIPS2022}, we evaluated on held-out  \texttt{Town05} and with new weather conditions, which were excluded from training.

The results, averaged over three runs, are shown in Table~\ref{tab:genwmnet_in_carla_scores}, where the comparative baselines are taken from the table in MILE~\cite{MILE:Hu:NEURIPS2022}. (To ensure a fair comparison, we reproduced the numbers from MILE itself.)  Our table also includes four methods that were published after MILE. For two of these methods (TCP~\cite{wu2022tcp} and TF++~\cite{jaeger2023tfpp}), we grabbed the numbers from their respective papers; these results are shown in italics.  We also collected numbers from the tables in two additional papers (InterFuser~\cite{shao2022interfuser} and ReasonNet~\cite{shao2023reasonnet}) that show results on \texttt{Town05 Long}, which is the same held-out test map but without new weather conditions; these results are indicated with italics and an asterisk.

Our approach achieves better CARLA evaluation scores than prior single-camera state-of-the-art, and it even outperforms many methods that use multiple cameras and/or LiDAR. Note that our system was trained on just 50k episodes, compared with 242k episodes for MILE~\cite{MILE:Hu:NEURIPS2022}.

\subsection{Covariate Shift Experiments}

To test whether training with latent space generative models mitigate covariate shift, we compared our system against a behavior cloning (BC) method to imitate human driver actions. 
In this ablation study, we kept the same BC architecture as in Fig.~\ref{fig:genwm_architecture} during inference but made two changes during training:  1) The world model network (along with KL-divergence) was removed, and 2) Instead of stochastically sampling the posterior distribution of the state estimator, the deterministic mean was used.
These changes result in a BC system with state tracking over time (by history network).



We compared the two approaches in the closed-loop CARLA simulator using \texttt{Town02}. Results shown in 
Table~\ref{tab:genwmnet_vs_bc_metrics} demonstrate successful navigation of our system without drifting into undesirable states. In contrast, the policy trained via behavior cloning (BC) achieves worse scores across multiple metrics in the same simulator.\footnote{Adding stochastic sampling to BC improves results slightly, but still leads to drifting and collisions. In our experience, training BC on an order of magnitude more data does not resolve covariate shift.} 
The significant difference demonstrates the effectiveness of our latent space generative world model in mitigating covariate shift.
Note, however, that our BC implementation still achieves good performance, primarily due to the robust visual features obtained by DINOv2.

\begin{table}
\caption{Results of our system vs. behavior cloning (BC) for \texttt{Town02} only.}
\label{tab:genwmnet_vs_bc_metrics}
\begin{center}
\begin{minipage}[t]{.95\linewidth}
\renewcommand{\arraystretch}{1.25}
\begin{tabular}{lccc}
\toprule
\textbf{Metric} & \textbf{GenWorldModel} & \textbf{Behavior Cloning} \\ 
& \textbf{(ours)}  & \textbf{(BC)} \\
\midrule
Route comp., no crash $\uparrow$  & \textbf{80.0\%} & $30.0$\%  \\
Route comp., km $\uparrow$ & \textbf{0.9} & $0.8$ \\
Driving Score $\uparrow$ & \textbf{80.0} & $70.0$  \\ 
Reward $\uparrow$ & \textbf{3500.4} & $2936.8$  \\
\bottomrule
\end{tabular}
\end{minipage}
\end{center}
\end{table}

\begin{figure}
    \centering
    \mbox{
      \parbox{0.99\columnwidth}{
        \centering
        \includegraphics[width=0.98\columnwidth]{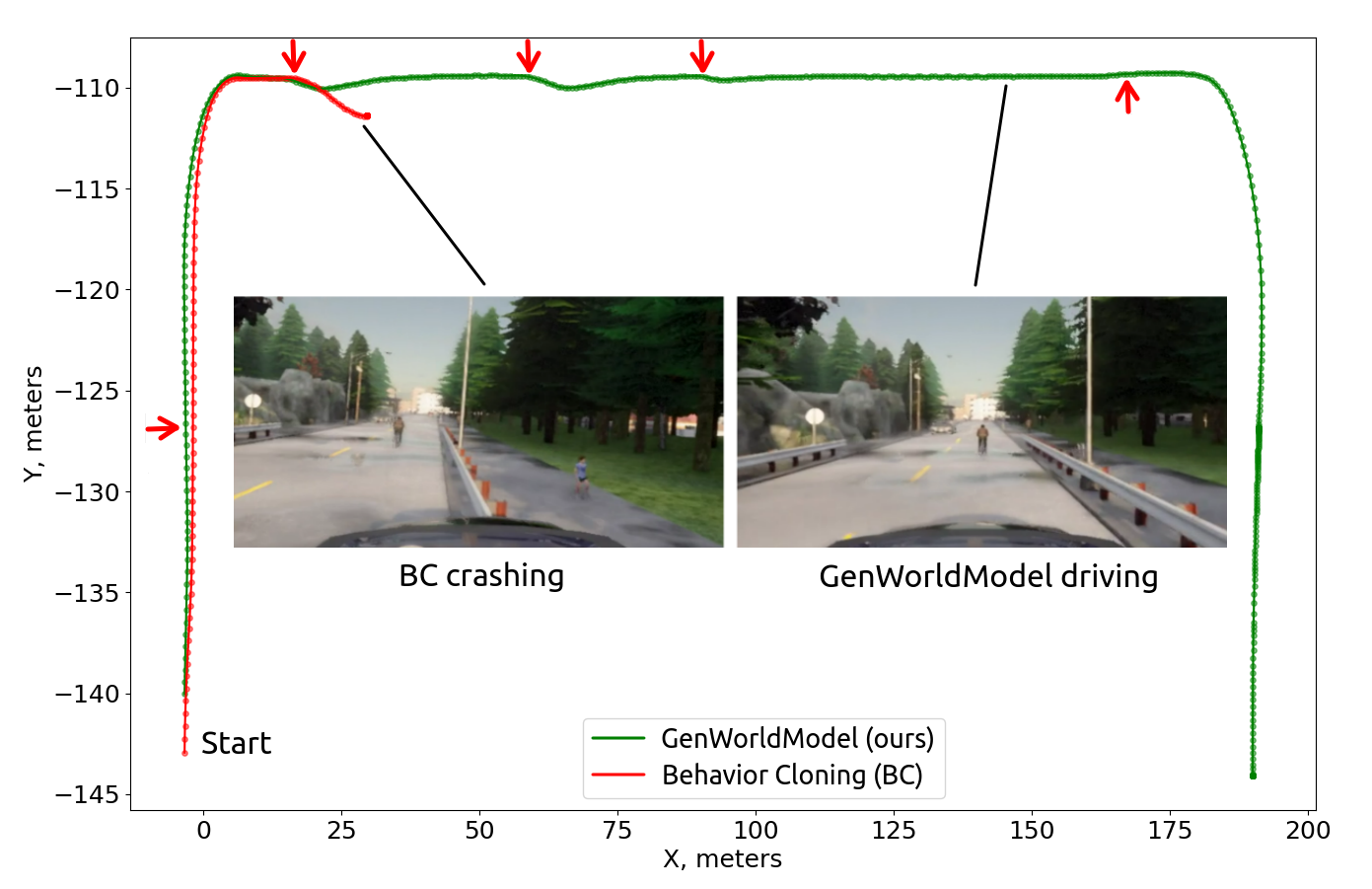}
        }
    }
    \caption{Top-down qualitative comparison of policies driving along a route in CARLA~\cite{dosovitskiy2017carla}, while steering control disturbances were applied (red arrows).  Behavior cloning was not able to recover from the second disturbance, hitting the guardrail and terminating early. Our system successfully recovered and completed the route.}     \label{fig:genwm_vs_bc_disturbance_withmarks}
\end{figure}

\begin{figure*}[t]
    \centering
    \mbox{
        \begin{minipage}[b]{0.3\linewidth}
            \centering            \includegraphics[width=\linewidth]{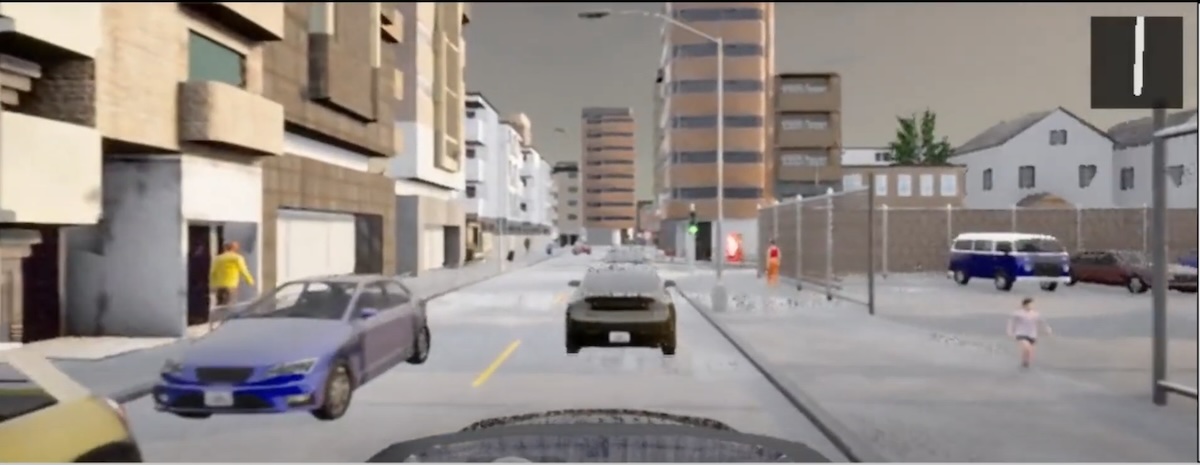}
        \end{minipage}
        \begin{minipage}[b]{0.3\linewidth}
            \centering            \includegraphics[width=\linewidth]{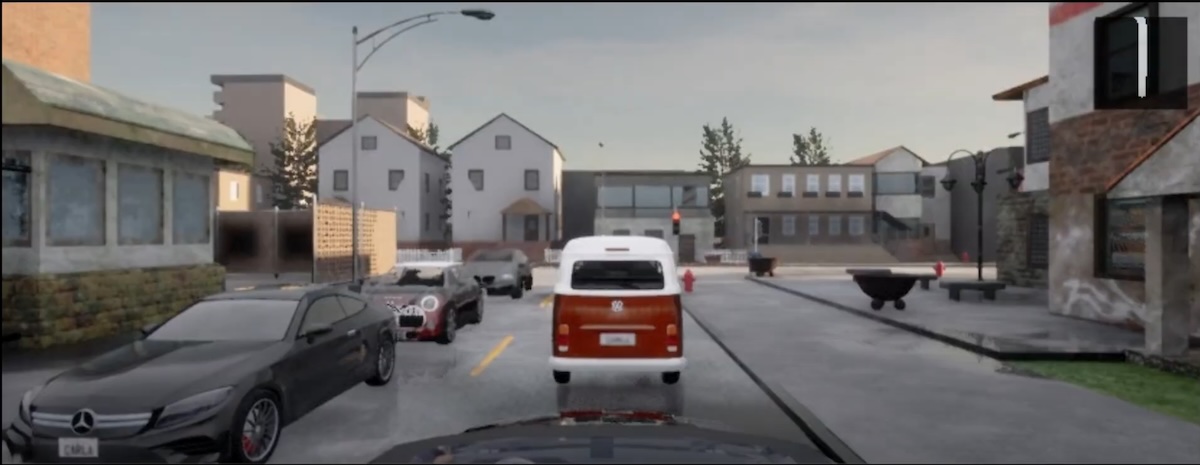}
        \end{minipage}
        \begin{minipage}[b]{0.3\linewidth}
            \centering            \includegraphics[width=\linewidth]{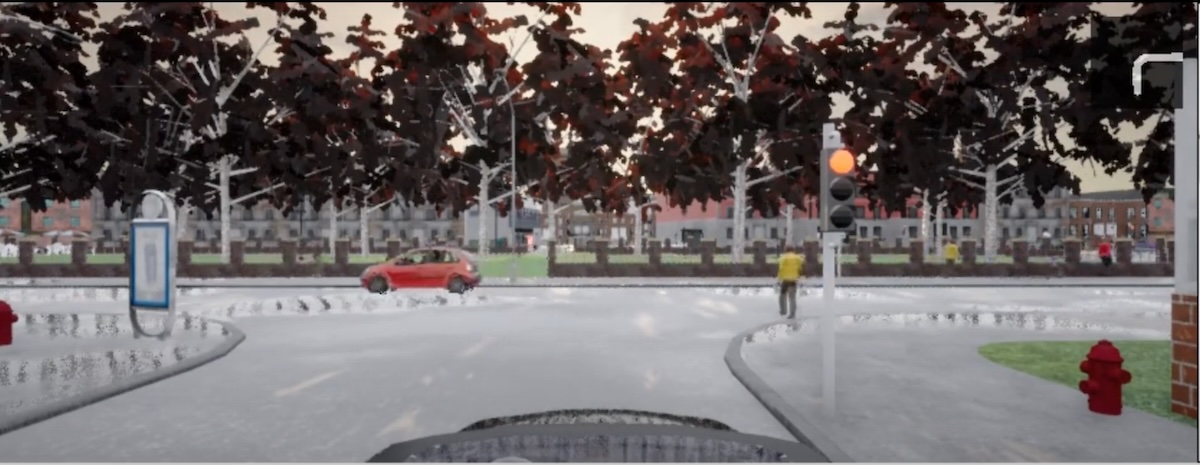}
        \end{minipage}
    }
    CARLA simulator \\ \hspace{1ex} \tiny \\ \normalsize 
    \mbox{
        \begin{minipage}[b]{0.3\linewidth}
            \centering
            \includegraphics[width=\linewidth]{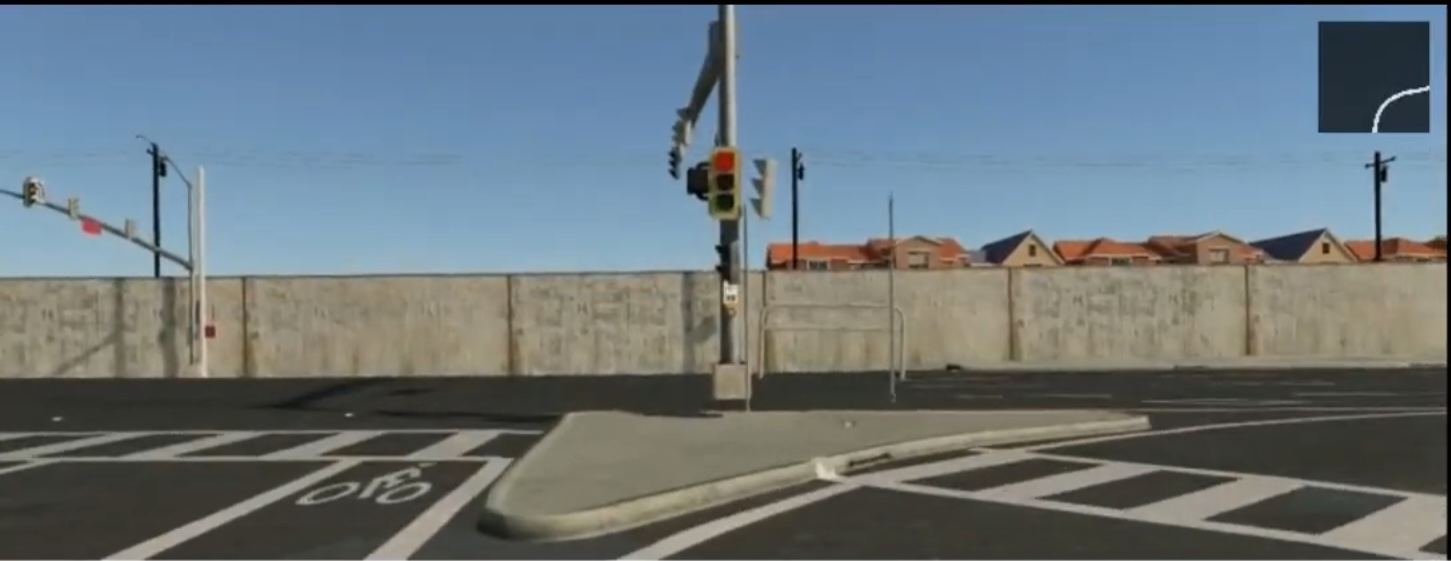}
        \end{minipage}
        \begin{minipage}[b]{0.3\linewidth}
            \centering
            \includegraphics[width=\linewidth]{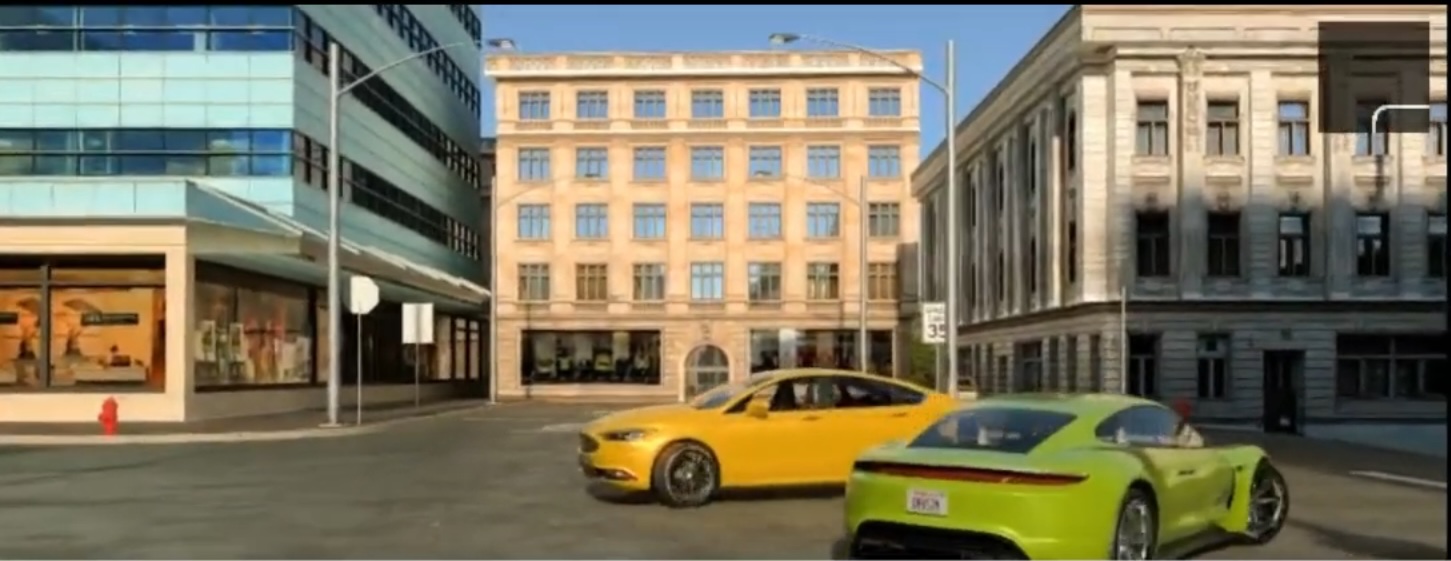}
        \end{minipage}
        \begin{minipage}[b]{0.3\linewidth}
            \centering
            \includegraphics[width=\linewidth]{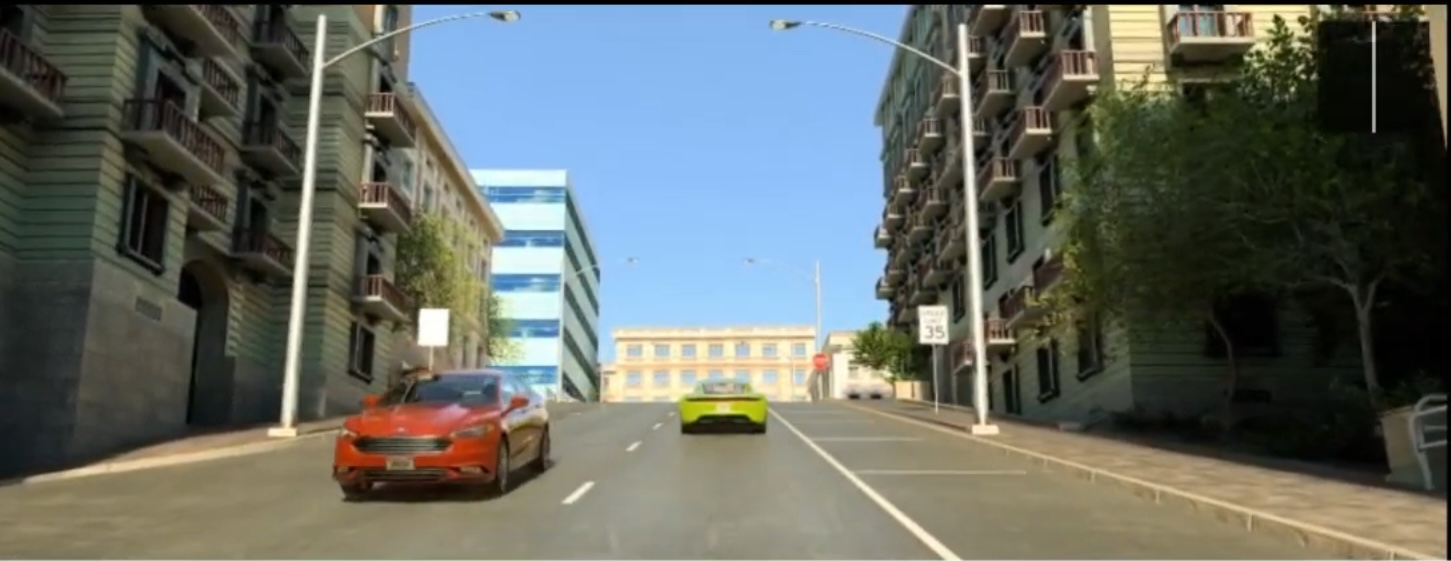}
        \end{minipage}
    }
    NVIDIA DRIVE~Sim simulator
    \caption{Our end-to-end system driving in the CARLA (top) and NVIDIA DRIVE~Sim (bottom) simulators.}
    \label{fig:carla_ds_driving}
\end{figure*}

\begin{figure}
    \centering
    \includegraphics[width=0.98\linewidth]{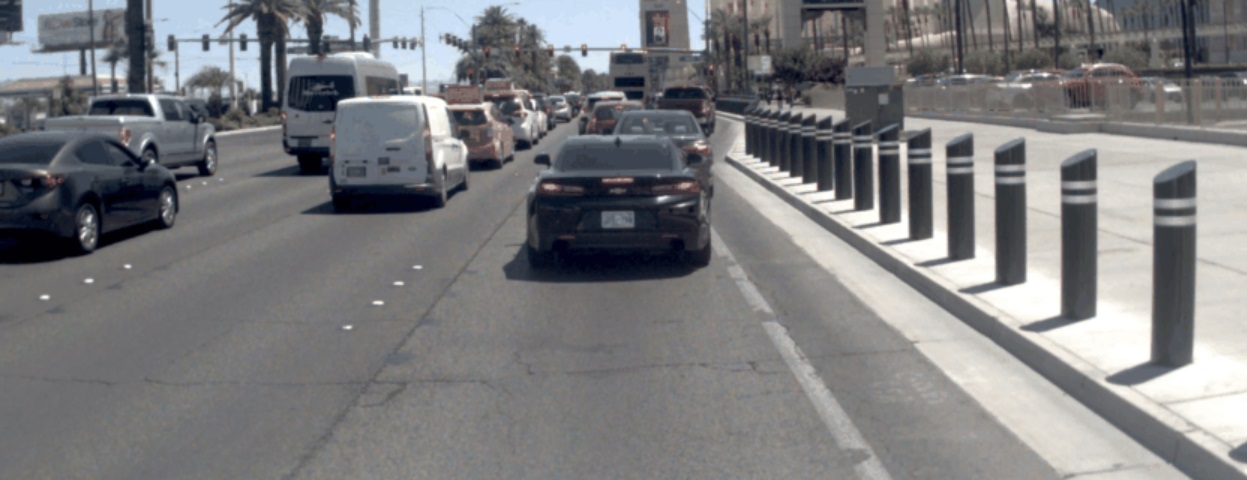}
    Input RGB image \\ \hspace{1ex} \tiny \\ \normalsize 
    \includegraphics[width=0.98\linewidth]{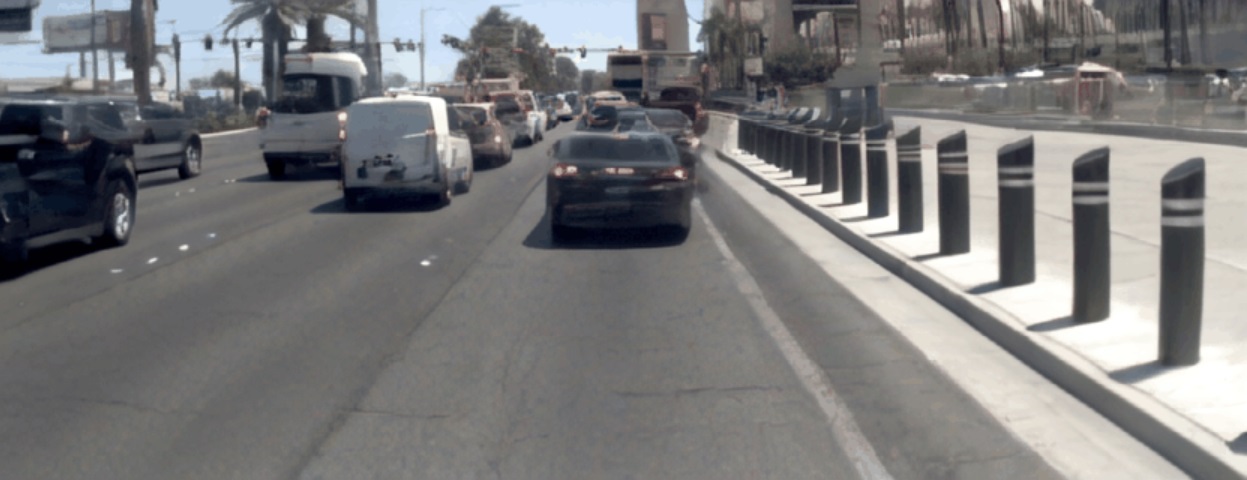}
    Generated image
    \caption{Original image (top) and reconstructed image (bottom) from our latent diffusion-based decoder. The latent representation faithfully captures much of the scene, although artifacts are visible (vehicles on the left).}
    \label{fig:diffusion_decode}
\end{figure}

\subsection{Disturbance Experiments}


As a continuation of the ablation study, we also conducted qualitative disturbance experiments in the closed-loop CARLA simulator \cite{dosovitskiy2017carla}.  Disturbances were introduced by manually overriding the driving policy for 300 ms at a time (3 consecutive frames at 10~Hz) with a \SI{30}{\degree} steering command to turn either right or left, then allowing the driving policy to resume control. 

Fig.~\ref{fig:genwm_vs_bc_disturbance_withmarks} shows an example of our method versus the behavior cloning (BC) policy described above, on a route involving two 90-degree right turns.  The BC policy was unable to recover from the second disturbance, causing the vehicle to crash into the guardrail.  Our policy, on the other hand, successfully navigated the route despite the many disturbances.
In other examples, we observed that our driving policy was able to recover to normal driving, even when the car was pushed onto the sidewalk or into oncoming traffic.

\subsection{DRIVE~Sim Experiments}

We also tested our model in NVIDIA DRIVE~Sim, which is more photo-realistic and physically realistic than CARLA (Fig.~\ref{fig:carla_ds_driving}). We trained our system using a 450-hour internal real driving dataset collected by multiple vehicles in our autonomous vehicle fleet in several geographic locations, consisting of 1.2-second expert driving episodes with 12 frames sampled at \SI{10}{\hertz}, amounting to 1.3 million episodes. The training process was the same as that used in the CARLA experiments. The system trained for 15 days on 64 NVIDIA A100 GPUs. We then deployed the system in closed-loop NVIDIA DRIVE~Sim, evaluating its performance across various scenarios and testing its recovery ability by artificially introducing a control lag of approximately \SI{250}{\milli\second}. The driving policy demonstrated robust performance and was able to recover effectively. Qualitative results can be seen in supplemental video material. Our model runs in near real time (10Hz) on NVIDIA Orin AGX embedded GPU.

\subsection{Perspective View RGB Decoding}

Recall from Eq.~\eqref{eq:one} that the system reconstructs the RGB perspective image $x_t$ from the latent space.
While StyleGAN decoder works well for simulated image data as shown in \cite{MILE:Hu:NEURIPS2022}, we found that the generated images are of poor quality and not temporally consistent when applied to more complex real data from the nuPlan dataset~\cite{nuplan}.

In contrast, our latent diffusion video decoder is able to generate high quality, temporally consistent $832 \times 320$ RGB frames that match the general scene setup of the original image frames used to initialize episodes.  A typical reconstructed image is shown in Fig.~\ref{fig:diffusion_decode}.

\newcommand\plotimagewidth{45mm}

\section{CONCLUSION}
In this work, we have shown that a driving policy trained with a latent generative world model mitigates the imitation learning covariate shift problem. We introduced a novel multi-view transformer-based perception encoder using self-supervised pre-training and DINOv2. We showed qualitative and quantitative results on CARLA~Sim, and qualitative results on NVIDIA DRIVE~Sim. Future work includes enhancing the world model predictor by adopting architectures based selective state space models like Mamba \cite{gu2023mamba}, investigating whether training with the world model reduces the amount of data required for neural policies, and exploring the use of reinforcement learning to handle long-tail scenarios.


\section*{Acknowledgment}
We thank Arthur Przech, Xiaolin Lin, George Hines, Ephrem Chemali, Ankit Gupta, Seth Robert Piezas, Kayley Ting, Hai Loc Lu, Ryan Holben and Yashraj Narang for contributions and feedback.

\bibliographystyle{IEEEtran}
\bibliography{WorldModel}

\end{document}